%% file: main.tex
\documentclass[10pt,twocolumn,letterpaper]{article}

\usepackage{cvpr}

\usepackage{graphicx}
\usepackage{amsmath}
\usepackage{amssymb}
\usepackage{booktabs}
\usepackage{bm}

\newcommand{\hr}[1]{\textcolor{cyan}{#1}}

\newcommand{\comment}[1]{}

\usepackage[pagebackref,breaklinks,colorlinks]{hyperref}

\usepackage[capitalize]{cleveref}
\crefname{section}{Sec.}{Secs.}
\Crefname{section}{Section}{Sections}
\Crefname{table}{Table}{Tables}
\crefname{table}{Tab.}{Tabs.}

\def\cvprPaperID{4396} %
\def\confName{CVPR}
\def\confYear{2022}

\input{defs}

\begin{document}

\title{ElePose: Unsupervised 3D Human Pose Estimation by Predicting Camera Elevation and Learning Normalizing Flows on 2D Poses}

\author{Bastian Wandt, James J. Little, and Helge Rhodin\\
University of British Columbia\\
{\tt\small wandt@cs.ubc.ca}
}
\maketitle

\begin{abstract}
Human pose estimation from single images is a challenging problem that is typically solved by supervised learning.
Unfortunately, labeled training data does not yet exist for many human activities since 3D annotation requires dedicated motion capture systems.
Therefore, we propose an unsupervised approach that learns to predict a 3D human pose from a single image while only being trained with 2D pose data, which can be crowd-sourced and is already widely available.
To this end, we estimate the 3D pose that is most likely over random projections, with the likelihood estimated using normalizing flows on 2D poses.
While previous work requires strong priors on camera rotations in the training data set, we learn the distribution of camera angles which significantly improves the performance.
Another part of our contribution is to stabilize training with normalizing flows on high-dimensional 3D pose data by first projecting the 2D poses to a linear subspace. %
We outperform the state-of-the-art unsupervised human pose estimation methods on the benchmark datasets Human3.6M and MPI-INF-3DHP in many metrics.
\end{abstract}

\section{Introduction}
Human pose estimation from single images is an ongoing research topic with many applications in medicine, sports, and human-computer interaction.
Tremendous improvements have been achieved in recent years via machine learning.
However, many recent approaches rely on a large amount of data used to train a 3D pose estimator in a supervised fashion.
Unfortunately, such training data is hard to record and rarely available for specialized domains.
For this reason, recent work focuses on reducing the amount of labeled data by using weak supervision in the form of unpaired 2D-3D examples, sparse supervision with a small amount of labeled 3D data, or multi-view setups during training.
In contrast, we propose a method that is trained only from 2D data, which is easy to annotate by clicking visible keypoints in readily available images and thereby alleviates the 3D labelling and multi-view capture steps required by weakly- and fully-supervised approaches.

\begin{figure}[t]
	\centering
	\includegraphics[width=0.47\textwidth]{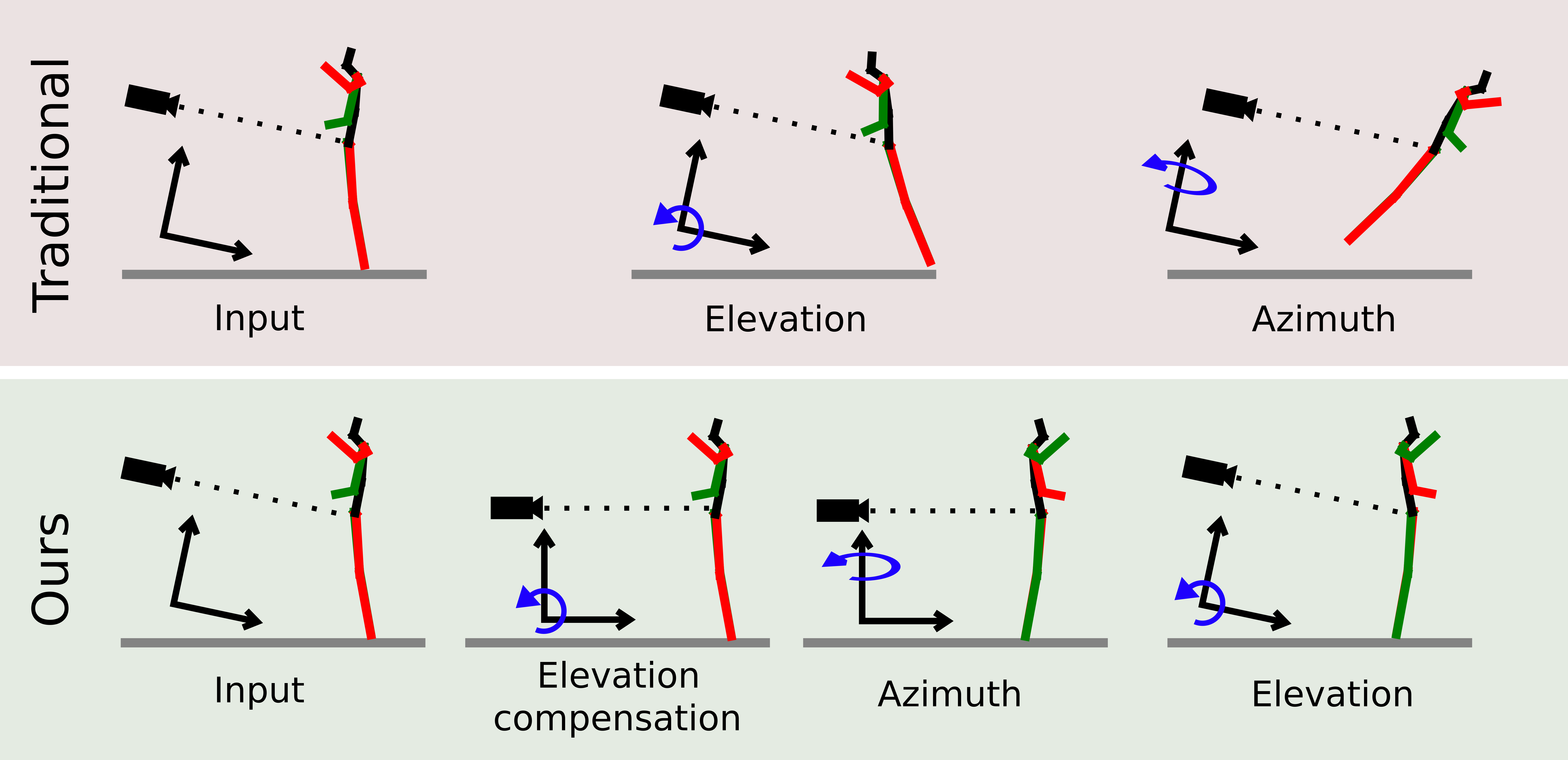}
	\caption{The consecutive steps performed by traditional methods and by our approach are shown from left to right. The commonly used prior distribution of the randomly sampled elevation angle leads to errors if it does not exactly match the distribution in the training dataset. ElePose solves this problem by learning this distribution and compensating for it before applying other transformations which significantly increases the performance of our method.}
	\label{fig:teaser}
\end{figure}

Given observations of 2D human joints in monocular imagery, we train a neural network to recover the depth ---the \emph{missing} third coordinate of the 3D human pose.
With the same goals, Chen et al. \cite{chen2019unsupervised} and Yu et al. \cite{Yu_2021_ICCV} train a 3D pose estimator in an adversarial setting \cite{Goodfellow2014}.
Their generator predicts a 3D pose that is randomly rotated and projected to a virtual camera which is fed into an adversarial network over 'fake' projected 3D poses and the 'real' 2D pose distribution. 
The idea is that, for a correct 3D pose prediction, the rotated and projected 2D pose should also come from the distribution of 2D training poses.
However, the predicted 3D pose is rotated randomly over a fixed prior distribution defined relative to the camera coordinate system.
It is a reasonable assumption when the camera is close to parallel to the ground plane.
However, even for small elevation angles and even when modeling variation
by sampling from a predefined Gaussian distribution, this leads to random projections that cannot be found in the training data as shown in Fig.~\ref{fig:teaser}.
 
We build upon this concept and
improve the handling of varying camera angles. %
Our core contribution is to
train a network that predicts the elevation for every 2D input.
After correcting for the predicted elevation, 3D reconstructions are upright such that rotating around the $y$-coordinate corresponds to rotating around the up-direction and uniformly sampling from all possible azimuth angles is meaningful as human poses are generally symmetric around the direction of gravity and the ground normal.
While camera angles have been estimated in supervised and weakly supervised settings \cite{WanRos2019a,Habibie_2019_CVPR,wandt2021canonpose,Kocabas2021SPEC} we do it for the monocular case and without supervision.

The approach of projecting to random virtual cameras requires to know the distributions of camera poses. Tailored to this, we propose a method for estimating the distribution of elevation angles from multiple point estimates,
which further
improves the performance of our model.

\comment{
\hr{Because the GAN training strategy utilized by prior work} is known to be sensitive to hyperparameters and can suffer from mode collapse, posterior collapse, and vanishing gradients~\cite{Kobyzev2020nfreview}, 
we propose to instead apply}

Another major change compared to previous work is that we use normalizing flows
to learn a prior distribution over 2D poses, which is subsequently used to infer the most likely 3D pose.
By contrast to GANs, which at best give a surrogate to the likelihood of an outcome with the discriminator response, our probabilistic formulation via normalizing flows naturally gives a likelihood for a predicted pose during inference time.
Besides gaining a significant improvement in accuracy and robustness over existing methods, our approach is also able to provide a measure for its performance, which is very valuable information in practical applications.

We overcome several technical challenges to make training and inference tractable. 
First, the bijectivity of normalizing flows is a useful property, which enables them to avoid mode collapse.
However, their construction restricts their input and output dimensions to be equal.
For high-dimensional data, such as human poses, this leads to non-optimal convergence and an incomplete latent space.
Second, the normalizing flow is still an approximation to the true pose distribution and can predict a high likelihood for poses that are outside the original distribution.
Optimizing the depth estimation network to produce 3D poses with high likelihoods for their back projections causes convergence to non-optimal solutions.
To avoid this, we propose to first project the 2D poses to a lower-dimensional space given by a Principal Components Analysis (PCA) on the training data.
Additionally, we introduce a suitable prior for the relative bone lengths in the human body to predict anthropometrically valid 3D poses.

\paragraph{Ethics and general impact.} Building such an unsupervised approach for motion capture promises to be more inclusive to people and activities that are not well represented in current motion capture datasets. 
We will make the source code available.

Pose estimators could be abused for unwanted surveillance and our method could be used for motion pattern analysis.
However,
we believe this risk is low since it does not reconstruct any visual features.

\section{Related Work}
In this section we discuss recent 3D human pose estimation approaches, structured by the different types of supervision, and put our approach in context.

\paragraph{Full Supervision.}
Supervised approaches rely on large datasets that contain millions of images with corresponding 3D pose annotations.
Li et al. \cite{Li2014} were the first to apply CNNs to regress a 3D pose from image input directly.
They later improved their work \cite{Li2015} by a structured learning framework.
Others followed this image-to-3D approach \cite{Tekin2016BMVC,Park2016,Du2016,VNect_SIGGRAPH2017,Pavlakos2017,lcrnet2017,Tome_2017_CVPR,sun2017compositional,OriNet2018,sun2018integral,Hossain2018,zhou2019hemlets,Li_2020_CVPR,Xu_2020_CVPR,kocabas2019vibe}.
Typically, these end-to-end approaches achieve exceptional performance on similar image data.
However, they struggle to generalize to very different scenes.
To avoid the dependence on image data other approaches use a pretrained 2D joint detector \cite{chen20173d,Moreno_cvpr2017,fang2017learning,mpii3Dhp2017,pavlakos2018ordinal,XNect_SIGGRAPH2020}.
Martinez et al. \cite{martinez_2017_3dbaseline} train a neural network on 2D poses and corresponding 3D ground truth.
Due to its simplicity, it can be trained quickly for a large number of epochs leading to high accuracy, and serves as a baseline for many following approaches.
While effective, the major downside of all supervised methods is that they do not generalize well to images with unseen poses.

\paragraph{Weak Supervision.}
Weakly supervised approaches require only a small set of labeled 3D poses or unpaired 2D and 3D poses.
Several approaches assume unpaired 2D and 3D poses \cite{WanRos2019a,Wang_2019_ICCV,Kundu_2020_CVPR,zanfir20normalizing,chen20garnet,habekost20learning} and leverage available motion capture data and combine it with unknown 2D data.
To allow for in-the-wild pose estimation of datasets where no training data is available, a transfer learning approach is introduced by Mehta et al. \cite{mpii3Dhp2017} which was later improved in Mehta et al. \cite{VNect_SIGGRAPH2017} to achieve real-time performance.
Other work first learns an embedding of multi-view data which is then used to train a 3D pose estimator with a sparse set of labeled 3D poses.
Rhodin et al. \cite{rhodin2018learning,rhodin2018unsupervised} use multi-view images and known camera positions to learn a 3D pose embedding.
Others \cite{rhodin2019neural,Unsup3DPose,mitra2020multiview,viewinvariant3DPose} followed the same idea.
Compared to completely supervised approaches, these weakly supervised methods generalize and transfer better to unseen poses.
However, they struggle with poses that are very different from the labeled training set.

\begin{figure*}[h!]
	\centering
	\includegraphics[width=0.99\textwidth]{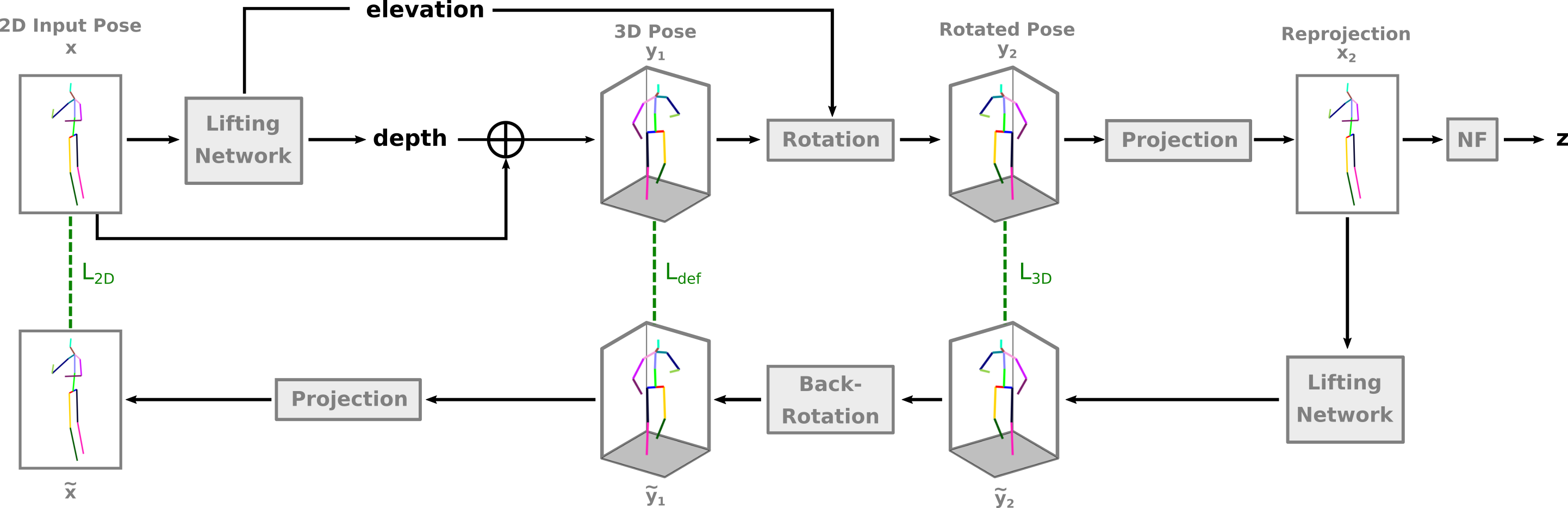}
	\caption{Overview of our approach. Given a normalized 2D input pose a lifting network predicts a depth for each joint coordinate which gives a 3D pose. Additionally it predicts the camera elevation in a parallel path. This 3D pose is randomly rotated and projected to 2D. The pretrained normalizing flow computes the negative log-likelihood which is used as a loss to train the lifting network.}
	\label{fig:pipeline}
\end{figure*}

\paragraph{Multi-view Supervision without 3D Data.}
Multi-view approaches only use information from multiple cameras without requiring any 3D data.
Rochette et al. \cite{rochette2019weakly} achieve similar performance as a comparable fully supervised approach.
However, they use a large number of cameras from different viewing angles, which limits the practical applicability.
Kocabas et al. \cite{kocabas2019epipolar} apply epipolar geometry to 2D poses from multiple views to compute a pseudo ground truth which is then used to train the 3D lifting network.
Iqbal et al. \cite{Iqbal_2020_CVPR} train an end-to-end network that refines the pre-trained 2D pose estimator during the self-supervised training.
Likewise, Wandt et al. \cite{wandt2021canonpose} reconstruct 3D poses in a canonical pose space that is consistent over all views.
While these multi-view approaches are a promising direction towards motion capture in the wild, they still require multiple temporally synchronized cameras for training.

\paragraph{Unsupervised.}
This section covers work that does not use any 3D data or additional views.
Our work also falls into this category.
Drover et al. \cite{drover2018can} propose an unsupervised learning approach to monocular human pose estimation.
They randomly project an estimated 3D pose back to 2D.
This 2D projection is then evaluated by a discriminator following adversarial training approaches.
However, they create an artificial 2D dataset from known ground-truth 3D poses.
Chen et al. \cite{chen2019unsupervised} extend \cite{drover2018can} with a cycle consistency loss that is computed by lifting the randomly projected 2D pose to 3D and inversing the previously defined random projection.
In contrast to Drover et al. \cite{drover2018can} they only use 2D data given by the dataset.
Yu et al. \cite{Yu_2021_ICCV} build upon \cite{chen2019unsupervised} and introduce a learnable scaling factor for the input 2D poses. None of them estimates the camera orientation and its distribution and we are also the first to apply normalizing flows to this setting.

\paragraph{Normalizing Flow.}

Informally, a normalizing flow is a tool to efficiently map distributions back and forth between two spaces. It applies to probability density estimation, which we use for the likelihood estimation of poses.

Let $\cZ\in\mathbb{R}^{N}$ be a known distribution (in our case a normal distribution) and $g$ be an invertible function $g(\vz)=\vx$, with $\vx\in \mathbb{R}^{N}$ as a vector representing the joints of a human pose\footnote{This could be either the 2D pose vector $\vx$ or its image in the PCA subspace.}.
With the change of variables formula the probability density function of $\vx$ is computed as 
\begin{equation}
    \label{eqn:changeofvariables}
    p_\cX(\vx) = p_\cZ(f(\vx))\left|\det\left(\frac{\partial f}{\partial \vx}\right)\right|,
\end{equation}
where $f$ is the inverse of $g$ and $\frac{\partial f}{\partial \vx}$ is the Jacobian of $f$.
That means given an invertible function $f$ the density of a 2D pose $\vx$ can be calculated by the product of the density of its projection $f(\vx)$ with the respective Jacobian determinant.
In our case $f$ is the trainable neural network proposed in \cite{Dinh_17_realnvp}. 
Details on the construction and training are given in the supplemental document.
Normalizing flows have been used to learn prior distributions of 3D human poses \cite{zanfir20normalizing,biggs2020multibodies,Xu2020GHUM,kolotouros2021prohmr} or to model ambiguities during the lifting step \cite{Wehrbein2021Probabilistic}.
However, they aim to build a probabilistic 3D model of a skeleton and therefore require 3D training data.
To the best of our knowledge, our approach is the first that uses normalizing flows to learn the prior distribution of 2D input data to infer the probability of a reconstructed 3D pose.

\section{Formulation}

Our goal is to train a neural network that given a 2D pose $\vx \in \mathbb{R}^{2J}$ recovers the 3D pose $\vy \in \mathbb{R}^{3J}$ with $J$ 3D joint positions. 
In our unsupervised setting, only a dataset of 2D poses is available for training. 
The general concept follows the assumptions of previous work in this domain, i.e., we assume that a 3D pose is plausible when viewing its 2D projections from multiple views. 
The difficulty lies in finding a plausibility measure for these 2D projections without multi-view data.
We propose to learn this measure via normalizing flows.
As a second major contribution we learn the camera angle distribution instead of predefining a dataset-dependent prior.
This not only makes our approach more flexible but also overcomes the problem of wrongly rotated poses 
resulting in significant improvement in performance.
All parts of our pipeline are visualized in Fig.~\ref{fig:pipeline} and explained in the following sections.

\paragraph{Lifting and Camera Model.}

Given 2D joint locations $\vx\in \mathbb{R}^{2\times J}$,
we introduce a lifting network that predicts for every joint $j$ the depth $\vw_j = \vd_j + D$ as the offset $\vd_j$ to a constant depth $D$. 
The full 3D pose is reconstructed based on the perspective unprojection
\begin{equation}
\vy_j = [\vu_j \vw_j, \vv_j \vw_j, \vw_j],
\end{equation}
where $\vu_j$ and $\vv_j$ are the horizontal and vertical joint positions in the image. 
It inverts the perspective projection operation
\begin{equation}
P(\vy_j ) = P([\vy^{(\text{x})}_{j},\vy^{(\text{y})}_{j},\vy^{(\text{z})}_{j}] ) = 
[\vy^{(\text{x})}_{j}/\vy^{(\text{z})}_{j}, \vy^{(\text{y})}_{j}/\vy^{(\text{z})}_{j}],
\label{eq:projection}
\end{equation}
with $[\vy^{(\text{x})}_{j},\vy^{(\text{y})}_{j},\vy^{(\text{z})}_{j}]$ as the 3D position of joint $j$.
To prevent ambiguous reconstructions with negative depth, $\vw_j$ is clipped to be larger than one.
Following previous work \cite{chen2019unsupervised, Yu_2021_ICCV} the depth $D$ is fixed to $D=10$, as perspective effects change little with depth,
and
each 2D pose $\vy$ is normalized by centering it at the root joint and dividing it by the mean length of the vector from the root joint to the head joint.
\paragraph{Reprojection to Virtual Cameras.}
We motivate our approach from multi-view camera setups, where the depth can be supervised by reprojection to the other views. 
Since no multi-view data is available in an unsupervised setting we assume a \emph{virtual} second view.
It requires rotating 3D poses centered at their root joint with $\vy_2 = \mR [\vy_1]^{3\times J}$, where $\mR\in \mathbb{R}^{3\times 3}$ is the rotation matrix from the original camera to a virtual camera and $[\vy_1]^{3\times J}$ the pose vector $\vy_1$ in the original camera coordinate system reshaped to a matrix with one of the $J$ 3D joint positions in each column. 
Typically, the rotation $\mR$ is randomly sampled from a predefined distribution $\cR$ \cite{chen2019unsupervised,Yu_2021_ICCV}.
However, in general, it is unknown and different for every dataset.
One of our core contributions is to learn this distribution instead of predefining it which we discuss in the following.
Using the same perspective camera model as for the lifting in Eq.~\ref{eq:projection} the 2D pose $\vx_2 = P(\vy_2)$ is computed by moving the predicted 3D pose with the predefined translation $D$ and dividing each joint by its depth.

\paragraph{Reprojection Likelihood}
In a multi-view setting the re-projection likelihood is typically a Gaussian with standard deviation $\sigma_r$, centered at the 2D projection $\vx_2$ of the 3D pose $\vy_2$ 
leading to a least squares loss when inferred using maximum likelihood or MAP.
Since multi-camera information is not available in an unsupervised setting there exists no corresponding 2D pose that can be matched to $\vx_2$.
While previous work \cite{chen2019unsupervised,Yu_2021_ICCV} tries to learn the distribution of plausible 2D poses with an adversarial approach we leverage normalizing flows for learning the probability density function of the 2D poses in the training dataset.
We define the reprojection likelihood by computing the likelihood of the latent variable $\vz$ in the latent space of a normalizing flow using Eq.~\ref{eqn:changeofvariables}.
In contrast to \cite{chen2019unsupervised,Yu_2021_ICCV}, this enables us to compute a likelihood for each reconstructed 3D pose which is very valuable information for downstream tasks.

In practice, we minimize the negative log-likelihood of Eq.~\ref{eqn:changeofvariables} which gives the normalizing flow loss
\begin{equation}
    \cL_{NF} = -\log(p_\cX(\vx))
    .
\end{equation}

\paragraph{Stabilized Normalizing Flows.}
We found that directly training the normalizing flow on 2D poses leads to non-optimal convergence during training of the lifting network.
We hypothesize that it is due to the high dimensionality of the input data which leads to a sparse latent space of the normalizing flow. 
That means the latent space contains poses that are not in the original distribution of 2D poses, although the normalizing flow assigns a high likelihood to them.
To mitigate this, instead of directly estimating the likelihood of a 2D pose we propose to first project the 2D pose to a low dimensional subspace.
Our subspace is determined by principal components analysis.
The projection to the subspace eliminates redundancies and noise from the data and, therefore, leads to a more stable training of the normalizing flow and subsequently the lifting network.

\paragraph{Camera Distribution and Elevation.}
The predicted 3D pose $\vy$ is rotated to a virtual view by randomly sampling $\mR \sim \cR$.
To achieve a reasonable 2D projection of the rotated 3D pose the distribution $\cR$ needs to be defined such that it matches the distribution of rotations present in the training data set.
In general, $\cR$ is unknown in an unsupervised setting.
However, there are reasonable priors for camera setups based on natural human behaviour while recording another human: 1) cameras are held horizontally, 2) since gravity defines a clear up-direction, cameras (or observed subjects) are mostly rotated around the azimuth axis, and 3) similar activities are recorded with similar but slightly varying elevation angles.
In terms of $\cR$ these three points mean that 1) there is negligible rotation around the optical axis, 2) a uniform prior over $360^\circ$ rotation around the azimuth axis is plausible, and 3) an unknown but restricted rotation around the elevation axis.
While the former two assumptions can be straightforwardly modeled the latter is commonly approximated by sampling the elevation angle from a uniform distribution in the interval $[-\pi/9, \pi/9]$ \cite{chen2019unsupervised,Yu_2021_ICCV}.
Unfortunately, this can lead to situations where a reconstructed person is strongly tilted towards the camera as visualized in Fig.~\ref{fig:teaser}.
This in turn results in backprojections that cannot be observed in the training set.
As a major contribution in this paper we propose to learn the distribution of the elevation angle $\cR_e$.
Since each 2D pose can have a unique elevation angle the lifting network is extended by a branch that predicts the elevation angle.
The resulting rotation matrix $\mR_e$ is used to rotate the predicted 3D pose $\vy$ to the direction of gravity by $\mR_e^T[\vy]^{3\times J}$.
This step alone already improves the predictions since it compensates for the formerly ignored elevation and therefore the azimuth rotation is correctly applied around the azimuth axis.

To further improve the results we additionally use the elevation predictions to predict the normal distribution of elevation angles in the dataset by calculating the mean $\mu_e$ and standard deviation $\sigma_e$ over all elevation angles in a batch such that 
\begin{equation}
    p(\cR_e)=\cN(\mu_e,\sigma_e).
\end{equation}
The rotation around the azimuth axis $\mR_a$ is randomly sampled from a uniform distribution in the interval $[-\pi, \pi]$.
To rotate the pose back in elevation direction a rotation $\tilde{\mR}_e$ for each sample in the batch is randomly sampled from the normal distribution $\cN(\mu_e, \sigma_e)$.
To allow for backpropagation through the sampling step we use the same reparametrization as for variational autoencoders, \ie 
\begin{equation}
    \tilde{\mR}_e \sim \mu_e + \sigma_e \cN(0, 1).
\end{equation}
The full rotation $\mR$ can now be written as 
\begin{equation}
    \mR=\mR_e^T \mR_a \tilde{\mR}_e.
\end{equation}
Our experiments show that our novel elevation angle estimation significantly improves results by approximately 15\% in PA-MPJPE and more than 22\% in MPJPE.

\paragraph{Skeleton likelihood.}
Human poses have several anthropometric properties defined by the kinematic chain of bones.
Most of these properties, such as bone lengths and joint angle limits, are unknown in an unsupervised setting.
However, \emph{relative} bone lengths are nearly constant across people \cite{Pietak2013ratios}.
For this reason, we calculate the relative bone lengths $b_k$ for the k-th bone divided by the mean length of all bones of a single pose.
We use a Gaussian prior with the mean at the pre-calculated relative bone length $\bar{b}_k$.
The density for the bone lengths prior is given by
\begin{equation}
\label{eqn:bone_lengths_prior}
p(b_1,b_2,\dots,b_K|\bar{b}_1,\bar{b}_2,\dots,\bar{b}_K) = \prod^K_{k=1} \cN(b_k | \bar{b}_k, \sigma_b)
,
\end{equation}
where $K$ is the number of bones. This forms a prior in terms of 3D pose $\vy$ and a likelihood, $p(\vx_1, \vd)$, of $\vx_1$ given a depth $\vd$ since a 3D pose is formed as a combination of observation and latent variable.
Practically, we define the loss $\cL_{bone}$ as the negative log-likelihood of Eq.~\ref{eqn:bone_lengths_prior}.
Note that our formulation imposes a soft constraint but does not fix bones to a predefined length.

\paragraph{Additional Losses.}
We additionally employ 3 losses similar to \cite{Yu_2021_ICCV}, namely the 3D lifting loss $\cL_{3D}$, the deformation loss $\cL_{def}$, and the 2D reprojection loss $\cL_{2D}$.
Figure~\ref{fig:pipeline} visualizes these three losses.
Since the 3D pose $\vy_2$ that produces the 2D pose $\vx_2$ is known the lifting network is applied again to $\vx_2$ to obtain the lifted pose $\tilde{\vy}_2$.
We define the traditional supervised $L_2$ loss
\begin{equation}
\label{eqn:loss3d}
    \cL_{3D} = \|\tilde{\vy}_2 - \vy_2\|_2
    .
\end{equation}
By rotating $\tilde{\vy}_2$ back to the original view we get a 3D pose $\tilde{\vy}_1=\mR^T \tilde{\vy}_2$ that should match $\vy_1$.
Yu \etal \cite{Yu_2021_ICCV} showed that instead of directly applying another $L_2$ loss on these two poses it is beneficial to consider the deformation between two poses at different time steps.
Since we do not assume any temporal data we define the same loss between two samples of a batch that could come from different people and sequences.
For the poses $\vy_1$ and $\tilde{\vy}_1$ at batch position $a$ and $b$ we define the time and pose independent deformation loss
\begin{equation}
    \cL_{def} = \|(\tilde{\vy}_1^{(a)} - \tilde{\vy}_1^{(b)}) - (\vy_1^{(a)} - \vy_1^{(b)})\|_2
    .
\end{equation}
Using the same perspective projection as before $\tilde{\vy}_1$ is projected to the 2D pose $\tilde{\vx}=P(\tilde{\vy}_1)$.
This gives the 2D back projection loss
\begin{equation}
    \cL_{2D} = \|\tilde{\vx} - \vx\|_1
    .
\end{equation}
Since the combination of these three terms has proven to be successful in \cite{Yu_2021_ICCV} we summarize them as our basis loss
\begin{equation}
    \cL_{base} = \cL_{3D} + \cL_{def} +\cL_{2D}
    .
\end{equation}

\comment{
When applying the lifting network again to $\vx_2$ it also predicts the elevation $\tilde{e}$.
Following the same argument as for the derivation of Eq.~\ref{eqn:loss3d} we %
use we use the sampled elevation $e$ to supervise the estimated which gives the elevation loss
\begin{equation}
    \cL_{ele} = \|\tilde{e} - e\|
    .
\end{equation}
The elevation loss further refines our results as shown in the ablation studies in Section~\ref{sec:ablations}.
}

\paragraph{Neural Network Structure.}
The lifting network is inspired by the MLP-based lifting network from Martinez et al. \cite{martinez_2017_3dbaseline} and consists of 3 residual blocks, each of which contains 2 fully connected layers with 1024 neurons followed by a leaky ReLU activation function.
The input is upscaled to a dimension of 1024 by a fully connected layer followed by a leaky ReLU activation function.
Downscaling to the dimension of the depth is performed by another fully connected layer without activation.
The elevation angle is predicted in a path parallel to the 3 residual blocks of the depth estimation network that has identical architecture.
The normalizing flow consists of 8 coupling blocks.
Each sub-network that predicts the affine transformations $s$ and $t$ contains 2 fully connected layers with 1024 neurons and ReLU activation functions.

\paragraph{Training Details.}
The normalizing flow is pretrained separately from the lifting network for 100 epochs with a batch size of 256 samples.
We use the Adam optimizer with an initial learning rate of $10^{-4}$ and a weight decay of $10^{-5}$.
For the pretraining of the normalizing flow we divided the learning rate by 10 after 10, 20, and 30 epochs.

The full loss function is
\begin{equation}
    \mathcal{L} = \mathcal{L}_{\mathrm{NF}} + 50\mathcal{L}_{\mathrm{bone}} +  \mathcal{L}_{\mathrm{base}} 
    .
\end{equation}
When training the lifting network we use an initial learning rate of $2\cdot 10^{-4}$ and an exponential scheduling with a decay of $0.95$ every epoch for a total number of 100 epochs.
Both, the pretraining of the normalizing flow and the training of the lifting network, take approximately 6 hours on an NVIDIA P100 Pascal.

\section{Experiments}
We perform experiments on the well-known benchmark datasets Human3.6M \cite{h36m_pami}, MPI-INF-3DHP \cite{mpii3Dhp2017} and 3DPW \cite{Marcard2018}.
For the Human3.6M dataset, we follow standard protocols and evaluate on every 64th frame of the test set.
\paragraph{Metrics}
\label{sec:metrics}
For the evaluation on Human3.6M we calculate the \textit{mean per joint position error} (MPJPE), i.e. the mean Euclidean distance between the reconstructed and the ground truth joint coordinates.
Since an unsupervised setting does not contain metric data we scale the reconstructed 3D pose to match ground truth, commonly known as N-MPJPE \cite{rhodin2018learning}.
The second common protocol first employs a Procrustes alignment (includes scaling) between the poses before calculating the MPJPE, also known as PA-MPJPE.
For 3DHP we report the \textit{Percentage of Correct Keypoints} (PCK) and the corresponding area under curve, scale-normalized as mentioned above, which we call \emph{N-PCK}.
It is the percentage of predicted joints that are within a distance of $150mm$ or lower to their corresponding ground truth joint.
Additionally, we evaluate the Correct Poses Score (CPS) recently proposed by Wandt et al. \cite{wandt2021canonpose}.
Unlike the PCK, the CPS classifies a pose as correct if all joints of the pose are correctly estimated.
To be independent of a threshold value, the CPS calculates the area under the curve in a range from $0mm$ to $300mm$.

\subsection{Results in Controlled Conditions}
\begin{figure*}
	\centering
	\includegraphics[width=0.99\textwidth]{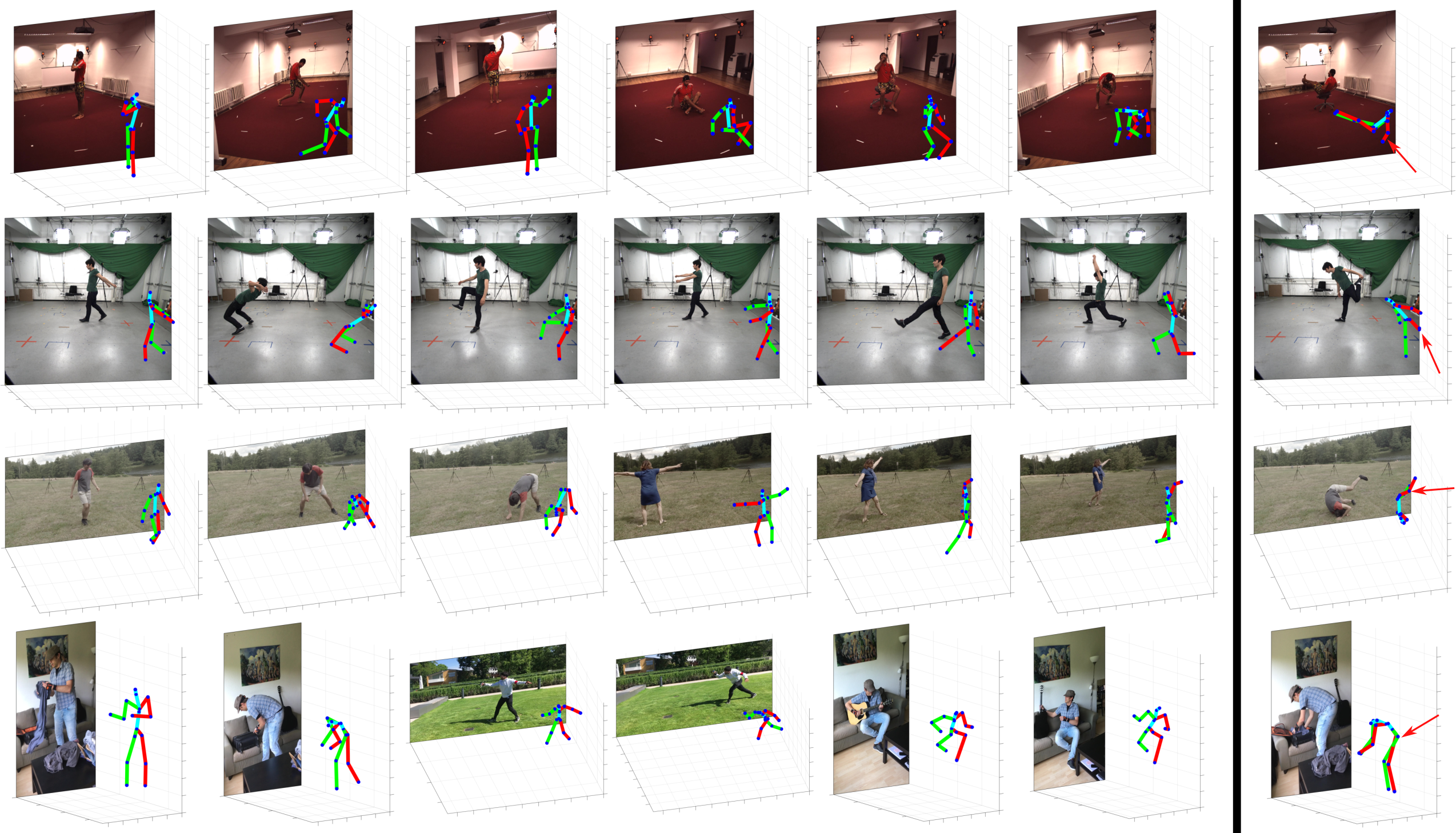}
	\caption{Subjective results of our method for the Human3.6M dataset (top row), the 3DHP dataset (middle two rows), and the 3DPW dataset (last row). The last column shows failure cases.}
	\label{fig:subjective}
\end{figure*}
To show the performance of our approach in a fair comparison to others, we start with using the 2D poses given by the dataset.
This allows for a fair comparison since it does not rely on the performance of pretrained 2D detectors that vary from method to method.
Table~\ref{tab:eval_h36m} presents results for the benchmark dataset Human3.6M with different types of supervision.
All shown results use the same 2D input data.
We outperform state-of-the-art \cite{Yu_2021_ICCV} in unsupervised pose estimation in PA-MPJPE by 12.6\%.
Notably, we even slightly improve on the PA-MPJPE of the fully supervised method of Martinez \etal \cite{martinez_2017_3dbaseline}.
We achieve comparable performance to weakly supervised methods and approaches using multi-view supervision in the N-MPJPE metric.
Note that \cite{Yu_2021_ICCV} uses a prior for the scale during training and therefore directly calculate the MPJPE.
However, we outperform them even when they apply the ground truth scale (PA-MPJPE: 39.7) during training.
For the Human3.6M dataset we obtain a CPS of 196.1.
Table~\ref{tab:eval_3dhp} and Table~\ref{tab:eval_3dpw} shows results for the MPI-INF-3DHP and 3DPW (in Train-Test-mode) datasets, respectively.
On the 3DHP dataset, we only found two other methods that use the 2D poses provided by the dataset and on the 3DPW we found no comparable method.
To create training data for the 3DPW dataset we reprojected the 3D skeletons to 2D.
This step is necessary because of the difference between the provided 2D and 3D data.
Note that for practical applications this step is not required.
The 3DPW dataset is particularly challenging since its training set comprises only data captured in-the-wild and additionally it is much smaller compared to the other two datasets.
The results show that our approach performs well even in challenging conditions.

\begin{table}[tp]
	\footnotesize
    \caption{Evaluation results for the Human3.6M dataset in $mm$. The bottom section, labeled with \textit{unsupervised}, shows comparable unsupervised methods. Best results are marked in bold. Numbers are taken from the respective papers. The star $^*$ indicates using a scale prior from the dataset. The MPJPE for \cite{martinez_2017_3dbaseline} is taken from \cite{Yu_2021_ICCV}.}
	\centering
	\resizebox{0.99\linewidth}{!}{
    \begin{tabular}{ l | l | c c c }
        Supervision & Method & PA-MPJPE$\downarrow$ & N-MPJPE$\downarrow$\\
        \hline
        full    & Martinez \cite{martinez_2017_3dbaseline}  & 37.1 & 45.5* \\
        \hline
        weak    %
                & 3D interpreter \cite{3dinterpreter2016}   & 88.6 & - \\
                & AIGN \cite{AIGN2017}                       & 79.0 & - \\
                & RepNet \cite{WanRos2019a}                 & 38.2  &  50.9  \\
                & Drover \cite{drover2018can}               & 38.2 & - \\
                & Kundu \cite{Kundu_2020_CVPR}              &  62.4 & -   \\
        \hline
        multi-view

            & EpipolarPose \cite{kocabas2019epipolar}  & 47.9 & 54.9 \\
            & Wandt \cite{wandt2021canonpose}             & 51.4 & 65.9   \\
        \hline
        unsupervised
            & Chen \cite{chen2019unsupervised}   &    58.0 & -   \\
            & \cite{chen2019unsupervised} reimplemented by \cite{Yu_2021_ICCV} & 46.0 & -\\
            & Yu \cite{Yu_2021_ICCV} (temporal) & 42.0 & 85.3$^*$\\
        \hline
            & Ours    &  \textbf{36.7} & 64.0  \\
    \end{tabular}
    }
    \label{tab:eval_h36m}
\end{table}

\begin{table}[tp]
	\footnotesize
    \caption{Evaluation results for the MPI-INF-3DHP dataset. The bottom section, labeled with \textit{unsupervised}, shows methods that can solve our setting. Numbers are taken from \cite{Yu_2021_ICCV}. A star $^*$ indicates an unknown normalization.}
	\centering
    \begin{tabular}{ l | l | c c c }
        Supervision & Method & PA-MPJPE$\downarrow$ & N-PCK$\uparrow$ & AUC$\uparrow$ \\
        \hline
        weak    
            & Kundu \cite{Kundu_2020_CVPR} & 93.9 & 84.6 & 60.8 \\
        \hline
        unsupervised & Yu \cite{Yu_2021_ICCV} & - & 86.2$^*$ & 51.7$^*$ \\
        \hline
          & Ours & \textbf{54.0} & 86.0 & 50.1  \\
    \end{tabular}
    \label{tab:eval_3dhp}
\end{table}

\begin{table}[tp]
	\footnotesize
    \caption{Evaluation results for the 3DPW dataset.}
	\centering
	\resizebox{0.98\linewidth}{!}{
    \begin{tabular}{ l | c c c c c }
         & PA-MPJPE$\downarrow$ & N-MPJPE$\downarrow$ & N-PCK$\uparrow$ & AUC$\uparrow$ & CPS$\uparrow$\\
        \hline
        Ours & 64.1 & 93.0 & 81.5 & 51.5 & 120.3 \\
    \end{tabular}
    }
    \label{tab:eval_3dpw}
\end{table}

Figure~\ref{fig:subjective} shows subjective results for both datasets.
On the left side are reconstructions with a low PA-MPJPE and visually plausible 3D skeletons.
Even poses that rarely occur in the training set are reconstructed correctly, e.g., sitting on the floor with crossed legs.
The right column shows occasional failure cases, with a PA-MPJPE over 200mm.
Typical failure cases are: limbs are rotated in the wrong direction (first and third row) and limb ordering (second and fourth row).

\subsection{Results in Practical Conditions}
In practice, where only images are available, we use an off-the-shelf 2D pose detector.
To be directly comparable to our closest competitor we use the same 2D detections produced by Cascaded Pyramid Networks \cite{chen2018cascaded} that are provided by the authors of VideoPose3D \cite{pavllo2019videopose3d,cpndetections}.
Table~\ref{tab:pose_predictions} shows our results when testing on the predicted 2D poses.
We outperform comparable unsupervised methods even though \cite{Chen_2019_CVPR,Yu_2021_ICCV} both use temporal information.

\begin{table}[tp]
	\footnotesize
    \caption{Results for unsupervised methods of the Human3.6M dataset when using 2D pose predictions. The star $^*$ indicates using a scale prior from the dataset instead of applying normalization via N-MPJPE at the inference stage.}
	\centering
    \begin{tabular}{l|c c c}
     & PA-MPJPE$\downarrow$ & N-MPJPE$\downarrow$ & CPS$\uparrow$ \\
    \hline
    Kundu \cite{Kundu_2020_CVPR}                & 62.4 & - & - \\
    Kundu \cite{Kundu2020Kinematic}             & 63.8 & - & - \\
    Chen \cite{chen2019unsupervised}     & 68.0 & - & - \\ %
    Yu \cite{Yu_2021_ICCV}                      & 52.3 & 92.4$^*$    & - \\
    \hline
    Ours                                        & \textbf{50.2}  & 74.4 & 165.3 \\
    \end{tabular}
    \label{tab:pose_predictions}
\end{table}

\subsection{Correlation Between Predicted 3D Poses and Likelihood of Projections}

\begin{figure}
    \centering
    \includegraphics[width=0.48\textwidth]{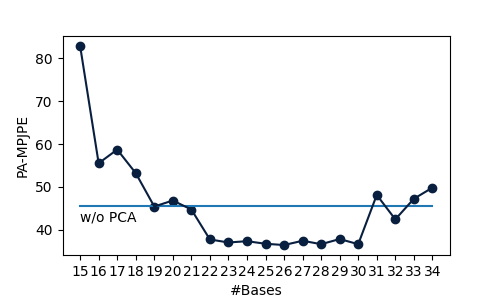}
    \caption{PA-MPJPE for different numbers of PCA bases. Between 22 and 30 PCA bases appears to be the ideal range.}
    \label{fig:bases}
\end{figure}
\begin{figure}
    \centering
    \includegraphics[width=0.48\textwidth]{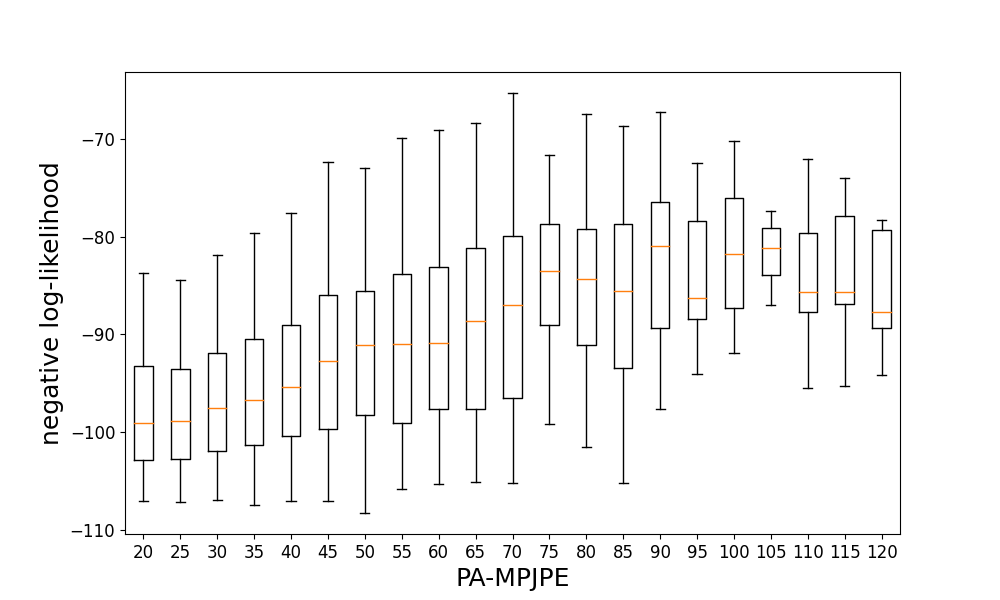}
    \caption{Correlation between the PA-MPJPE and the negative log-likelihood assigned to a set of projections of the predicted 3D pose. As desired, poses with a low 3D error have a low negative log-likelihood and vice versa for high 3D errors. Errors are given in millimeters.}
    \label{fig:correlation}
\end{figure}

A benefit of our novel normalizing flow formulation is that it can also be used during testing to evaluate the likelihood of the predicted 3D poses.
For practical applications, this can be an important value to assess the reliability of the predicted poses for downstream tasks.
We apply the normalizing flow to compute the negative log-likelihood of predicted poses.
For the reprojection log-likelihood we randomly sample 100 rotations from the distribution learned during the training stage which is then averaged over all rotations.
The elevation distribution is estimated over the whole training set.
A box plot of the results is shown in Fig.~\ref{fig:correlation}.
We show bins in 5mm steps from 20-120mm while the 120mm bin includes 120mm and above.
As expected, in many cases, the likelihood correlates with the 3D reconstruction error.

\subsection{Ablation Studies}
\label{sec:ablations}
\begin{table}[tp]
	\footnotesize
    \caption{Ablation study with different loss terms on the Human3.6M dataset.}
	\centering
    \begin{tabular}{ l | c | c }
    Configuration & PA-MPJPE & MPJPE\\
    \hline
    base ($\cL_{NF}$ + $\cL_{3D}$ + $\cL_{def}$ + $\cL_{2D}$)   & 77.9 & 135.0 \\
    base + $\cL_{bl}$                   & 48.1      & 83.8 \\
    base + $\cL_{bl}$ + elevation       & 45.5      & 73.9 \\
    base + $\cL_{bl}$ + PCA             & 43.1      & 83.8 \\
    \hline
    Ours (base + $\cL_{bl}$ + PCA + elevation)  & 36.7 & 64.0 \\
    \end{tabular}
    \label{tab:ablation}
\end{table}
We perform several experiments with different configurations of our approach on the Human3.6M dataset by training the lifting network with each of them separately.
Additionally, we train the normalizing flow directly on the 2D poses (i.e. no PCA).
The results in Table~\ref{tab:ablation} show that each of our contributions is important to achieve the best performance.
Note that using the bone lengths prior together with either PCA or elevation alone outperforms \cite{chen2019unsupervised} and its improved reimplementation by \cite{Yu_2021_ICCV}.
Without the PCA we achieve an PA-MPJPE of 45.5mm which shows the importance of projecting to the PCA space before training the normalizing flow.
Adding our novel elevation prediction improves the results by almost 15\%.

Since the PCA is an important part of achieving an acceptable performance, we evaluate the impact of the number of PCA bases.
Fig.~\ref{fig:bases} shows the results.
Projecting to a PCA space smaller than 15 bases removes important information from the reprojected 2D poses and results in errors above 100mm. 
For visualization purposes we only visualize the error for more than 15 bases.
The best performance lies between 22 and 30 bases that cover between 99.6\% and 99.9\% of the variance in the training set.
The increase at 31 bases shows that the normalizing flow struggles to learn the probability density when the input dimension is too large.

\section{Limitations}
Our proposed method enables 3D human pose estimation with only 2D annotations and paves the way to a monocular motion capture system for all possible human activities.
The only requirement is a set of 2D annotations which can be obtained by crowd sourcing 2D joint annotations.
This is the main limitation of our approach.
More specifically, our method requires similar poses seen from different angles.
While we compensate for one of these aspects, the elevation angle, natural assumptions on the shape of the distribution of azimuth angles are hard or even impossible to make.
Additionally, poses that appear visually correct from all angles can still be implausible in 3D space which is a general problem in monocular human pose estimation.
In future work we plan to mitigate these problems by learning 3D pose priors and conditional distributions over full camera rotations jointly.

\section{Conclusion}
Human pose estimation from single images is a challenging problem, especially for activities where no 3D annotations exist. 
To this end, we propose an unsupervised approach that learns a 3D lifting only from 2D annotations.
While previous approaches utilize a predefined prior on the camera distribution we find that learning this distribution significantly improves the results.
We formulate the 3D human pose estimation problem as a maximum likelihood estimation over random projections of the 3D pose.
Normalizing flows appear to be an ideal tool.
They not only learn a well-defined prior distribution of 2D poses but also allow us to calculate the likelihood of reconstructed 3D pose at test time which provides valuable information.
Since we observed that directly using the normalizing flow as prior leads to unstable training of the lifting network we additionally propose to first project the 2D poses to a low dimensional subspace.

\newpage

{\small
\bibliographystyle{ieee_fullname}
\bibliography{bibliography}
}

\end{document}

%% file: defs.tex
\newcommand{\mI}{\mathbf{I}}

\newcommand{\mR}{\mathbf{R}}

\newcommand{\cL}{\mathcal L}

\newcommand{\cN}{\mathcal N}

\newcommand{\cR}{\mathcal R}

\newcommand{\cX}{\mathcal X}

\newcommand{\cZ}{\mathcal Z}

\newcommand{\vd}{\mathbf{d}}

\newcommand{\vu}{\mathbf{u}}
\newcommand{\vv}{\mathbf{v}}
\newcommand{\vw}{\mathbf{w}}
\newcommand{\vx}{\mathbf{x}}
\newcommand{\vy}{\mathbf{y}}
\newcommand{\vz}{\mathbf{z}}